\documentclass[10pt,twocolumn,letterpaper]{article}

\usepackage{cvpr}              

\usepackage{graphicx}
\usepackage{amsmath}
\usepackage{amssymb}
\usepackage{booktabs}
\usepackage{multicol}
\usepackage{multirow}
\usepackage{enumitem}
\usepackage{threeparttable}

\usepackage[pagebackref,breaklinks,colorlinks]{hyperref}

\usepackage[capitalize]{cleveref}
\crefname{section}{Sec.}{Secs.}
\Crefname{section}{Section}{Sections}
\Crefname{table}{Table}{Tables}
\crefname{table}{Tab.}{Tabs.}

\def\cvprPaperID{7931} 
\def\confName{CVPR}
\def\confYear{2022}

\begin{document}

\title{Adaptive Background Matting Using Background Matching}

\author{Jinlin Liu\\
{\tt\small liujl09@live.com}
}
\maketitle

\begin{abstract}
Due to the difficulty of solving the matting problem, lots of methods use some kinds of assistance to acquire high quality alpha matte. Green screen matting methods rely on physical equipment. Trimap-based methods take manual interactions as external input. Background-based methods require a pre-captured, static background. The methods are not flexible and convenient enough to use widely. Trimap-free methods are flexible but not stable in complicated video applications. To be stable and flexible in real applications, we propose an adaptive background matting method. The user first captures their videos freely, moving the cameras. Then the user captures the background video afterwards, roughly covering the previous captured regions. We use dynamic background video instead of static background for accurate matting. The proposed method is convenient to use in any scenes as the static camera and background is no more the limitation. To achieve this goal, we use background matching network to find the best-matched background frame by frame from dynamic backgrounds. Then, robust semantic estimation network is used to estimate the coarse alpha matte. Finally, we crop and zoom the target region according to the coarse alpha matte, and estimate the final accurate alpha matte. In experiments, the proposed method is able to perform comparably against the state-of-the-art matting methods.
\end{abstract}


\begin{abstract}
Due to the difficulty of solving the matting problem, lots of methods use some kinds of assistance to acquire high quality alpha matte. Green screen matting methods rely on physical equipment. Trimap-based methods take manual interactions as external input. Background-based methods require a pre-captured, static background. The methods are not flexible and convenient enough to use widely. Trimap-free methods are flexible but not stable in complicated video applications. To be stable and flexible in real applications, we propose an adaptive background matting method. The user first captures their videos freely, moving the cameras. Then the user captures the background video afterwards, roughly covering the previous captured regions. We use dynamic background video instead of static background for accurate matting. The proposed method is convenient to use in any scenes as the static camera and background is no more the limitation. To achieve this goal, we use background matching network to find the best-matched background frame by frame from dynamic backgrounds. Then, robust semantic estimation network is used to estimate the coarse alpha matte. Finally, we crop and zoom the target region according to the coarse alpha matte, and estimate the final accurate alpha matte. In experiments, the proposed method is able to perform comparably against the state-of-the-art matting methods.
\end{abstract}

\section{Introduction}
\label{sec:intro}
The matting problem aims at estimating the precise alpha matte from the input images, which is of great value in lots of image and video editing applications. Considering the diversity of the real world scenes and the complex details of the target objects, it is challenging to estimate accurate alpha matte automatically. Mathematically, The matting formula is the combination of the foreground $F$, background $B$ and alpha matte $\alpha$. The given image $I$ can be calculated by the following method ~\cite{wang2008image}:
\begin{equation}
\label{eq:matting_def}
I_{z} = \alpha_{z}F_{z} + (1 - \alpha_{z})B_{z}\,, \quad \alpha_{z} \in [0,1]\,.
\end{equation}
where $I$ is the input image, $F$ is the foreground, $B$ is the background, $\alpha$ is the alpha matte, $z$ is the pixel of the input image $I$. In this formula, only the input image $I$ is observed, all other variables are unknown. As a result, solving such a problem is difficult and requires extra information. Lots of methods have been proposed to achieve this goal. We classify them to three categories: trimap-based, trimap-free and background-based methods.

Trimap-based methods~\cite{xu2017deep, forte2020f, lu2019indices, hou2019context, tang2019learning, Cai_2019_ICCV,sun2021semantic} refer to methods using trimap as external input. Trimap contains background, foreground and unknown regions, among which, only the unknown regions are meant to be estimated. In recent years, trimap-based methods are capable of producing very accurate alpha matte estimation results using deep networks. Whereas, the quality of trimap generated by manual interaction in real applications may vary significantly, which will lead to some unstable results. Also generating trimap is not easy and can be time-consuming especially for video kind applications. It is nearly not possible to generate trimap of consistent quality frame by frame.
In recent years, lots of trimap-free methods~\cite{liu2020boosting, qiao2020attention, Zhang_2019_CVPR, chen2018semantic, shen2016deep, zhu2017fast} are proposed to get rid of the limitations of trimap in real applications. The progress are very promising. Trimap-free methods require no extra input and estimate accurate alpha-matte by training on paired image to alpha matte datasets. Both accurate semantic information and the detailed edge should be estimated precisely by trimap-free methods. As a result, lots of training datasets are required by these methods so as to perform well in complicated real world applications~\cite{sengupta2020background}. In addition, for video kind applications, trimap-free methods usually suffer from inconsistent estimation results among frames.
Background-based methods~\cite{sengupta2020background, lin2021real} use extra background as input, and estimate alpha matte from the input image and the given background. These kinds of methods are very useful when applying to videos. The user only needs to capture the background first. Background-based methods are able to produce consist and stable results. In this scene, background-based methods are much easier to use then trimap-based methods and are more stable then trimap-free methods. Whereas, background-based methods require that the background should not change during applications, which means that these methods will not work when the background are dynamic or change frame by frame.

In summary, trimap-based methods require manual interaction and are nearly not operable in video applications. Though trimap-free methods are convenient, they require large amount of datasets and are difficult to generate stable results in complicated scenes. Background-based method are stable and easy to use in real applications, but they are limited to static backgrounds. Thus we propose a new adaptive background matting method, which is stable, easy to use regarding to all kinds of backgrounds.

\begin{figure}[t]
  \centering
  \resizebox{1\linewidth}{!}{
   \includegraphics{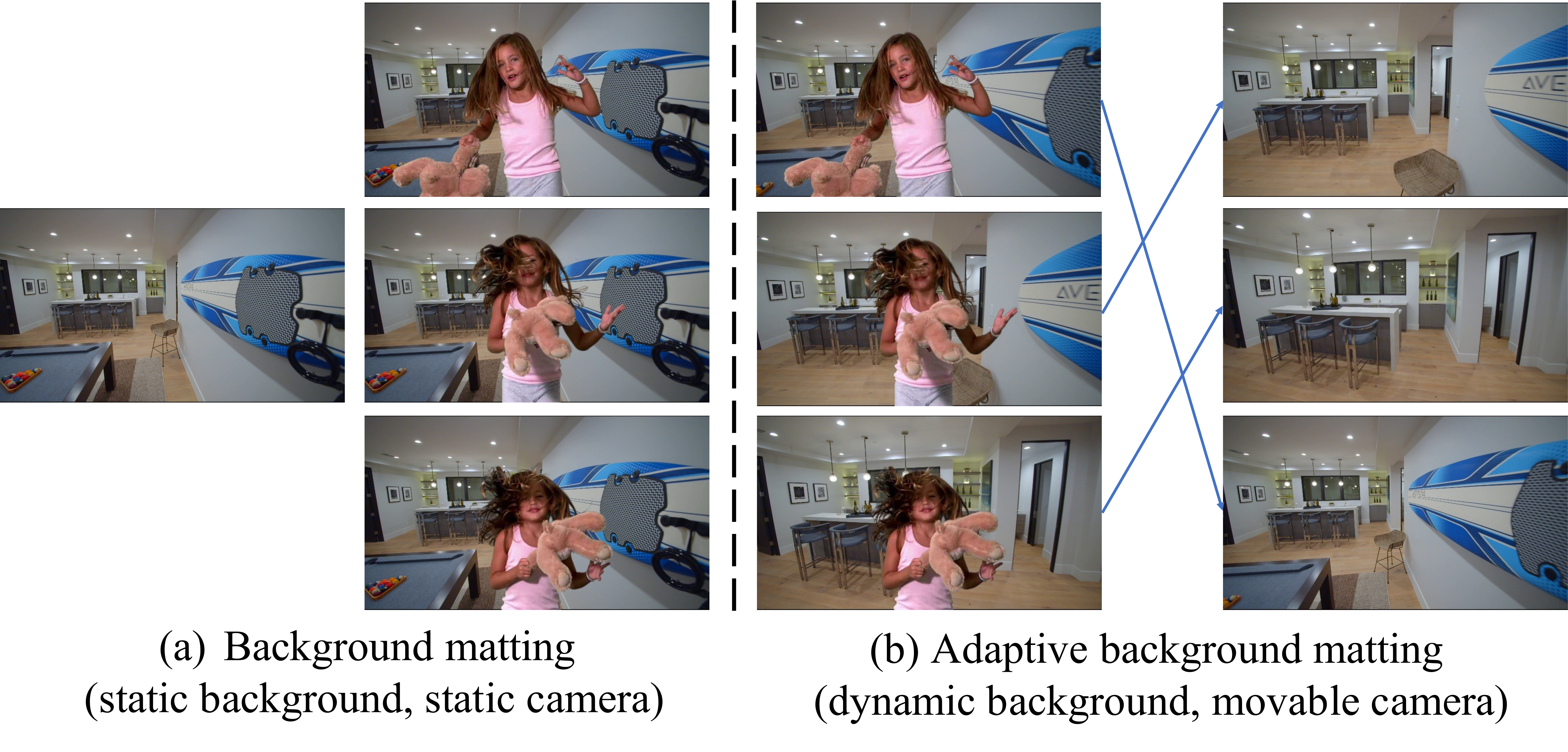} }
  \caption{The proposed adaptive background matting method uses dynamic background video rather than static image for accurate matting. The camera is not limited to be static, allowing the users to capture videos much more freely. The user first captures their videos freely, moving the cameras as they want. Then, the foreground person steps out the scene, the user captures the background according to the content of the captured video, roughly covering the background regions. The proposed method finds the best matched background for each frame automatically.}
  \label{fig: intro}
\end{figure}

In the proposed pipeline, the user first captures normal videos freely, moving the camera as they want. Then, the foreground person steps out the scene, the user captures the background according to the content of the captured video, roughly covering the background regions. Differently from previously preposed background-based methods, the proposed method does not require the background to be static (demonstrated in Figure~\ref{fig: intro}). Then we learn from the dynamic backgrounds for accurate matting. The proposed method contains three parts: background matching network (BMN), robust semantic estimation network (REN) and accurate alpha estimation network (AEN). In the first step, for each frame of the video, we identify the corresponding background frame in the background frames using the background matching network. After that, the robust semantic estimation network estimates the coarse alpha matte. With the coarse alpha matte, we crop the input to remove blank regions, zoom it to a large size, and use the alpha matte estimation network to estimate the final alpha matte.

The proposed method breaks the limitation of background-based method while keeping the advantages, and is much more easily to use in real world applications. In experiments, the proposed method is able to perform comparably against the state-of-the-art matting methods. The main contributions of this work are:

\begin{itemize}[topsep=0.5pt, itemsep=0.5pt, partopsep=0.5pt]
\item We propose a novel adaptive background matting method. To our best knowledge, this is the first method that uses dynamic background video for accurate matting.

\item We propose an effective pipeline to estimate alpha matte from the input video and dynamic background video. BMN is designed to find the best matched background from dynamic background frames. We estimate accurate alpha matte progressively from coarse to fine using REN and AEN, making the proposed method robust against possible background misregistration in reality.
\end{itemize}

\section{Related work}
\label{sec:related work}
\paragraph{Trimap-based methods.} Lots of traditional methods~\cite{chuang2001bayesian,feng2016cluster,gastal2010shared,he2011global,johnson2016sparse,karacan2015image,ruzon2000alpha,aksoy2018semantic,aksoy2017designing,bai2007geodesic,chen2013knn,grady2005random,levin2007closed,levin2008spectral,sun2004poisson} take trimap as external input. Most of these methods solve
~\ref{eq:matting_def} using some prior assumptions. More recently, deep learning networks are widely used~\cite{xu2017deep, forte2020f, lu2019indices, hou2019context, tang2019learning, Cai_2019_ICCV,sun2021semantic}. \cite{xu2017deep} first proposed deep image matting using convolutional networks. \cite{Hou_2019_ICCV} proposed to estimate the foreground and alpha matte at the same time. Further, \cite{sun2021semantic} incorporated semantic classification in their matting networks. Lots of excellent works have been proposed recently and the results are promissing. Whereas, the requirement of trimap limits the applications of these methods. For video kind applications, it is nearly impossible to generate trimap of consistent quality frame by frame.
\paragraph{Trimap-free methods.} Considering the limitation of trimap-based methods, lots of trimap-free methods have been developed. Trimap-free methods require large training dataset. Considering the difficulty of collecting matting datasets, most works focus on human-related matting~\cite{liu2020boosting,chen2018semantic, shen2016deep, zhu2017fast}. \cite{liu2020boosting} proposed to use coarse dataset to boost the performance of human matting. \cite{chen2018semantic} collected lots of human dataset and proposed an end-to-end human matting method. Trimap-free methods are convenient to use in real world applications, but they require large amount of datasets and are difficult to generate stable results in complicated scenes~\cite{sengupta2020background}.
\paragraph{Background-based methods.} The aid of pre-captured background helps improve the quality of matting significantly~\cite{sengupta2020background}. \cite{sengupta2020background} proposed to use context switching block to combine different inputs. In addition, they used unlabelled data to improve the performance. Recently, \cite{lin2021real} introduced a more efficient method to process high resolution input in real time. As pointed out by~\cite{sengupta2020background}, background-based method requires the background to be static with tiny camera motion. This is not in accord with the reality, as we usually move the camera to record more scenes.

\begin{figure*}[ht]
  \centering
  \resizebox{1\linewidth}{!}{
   \includegraphics{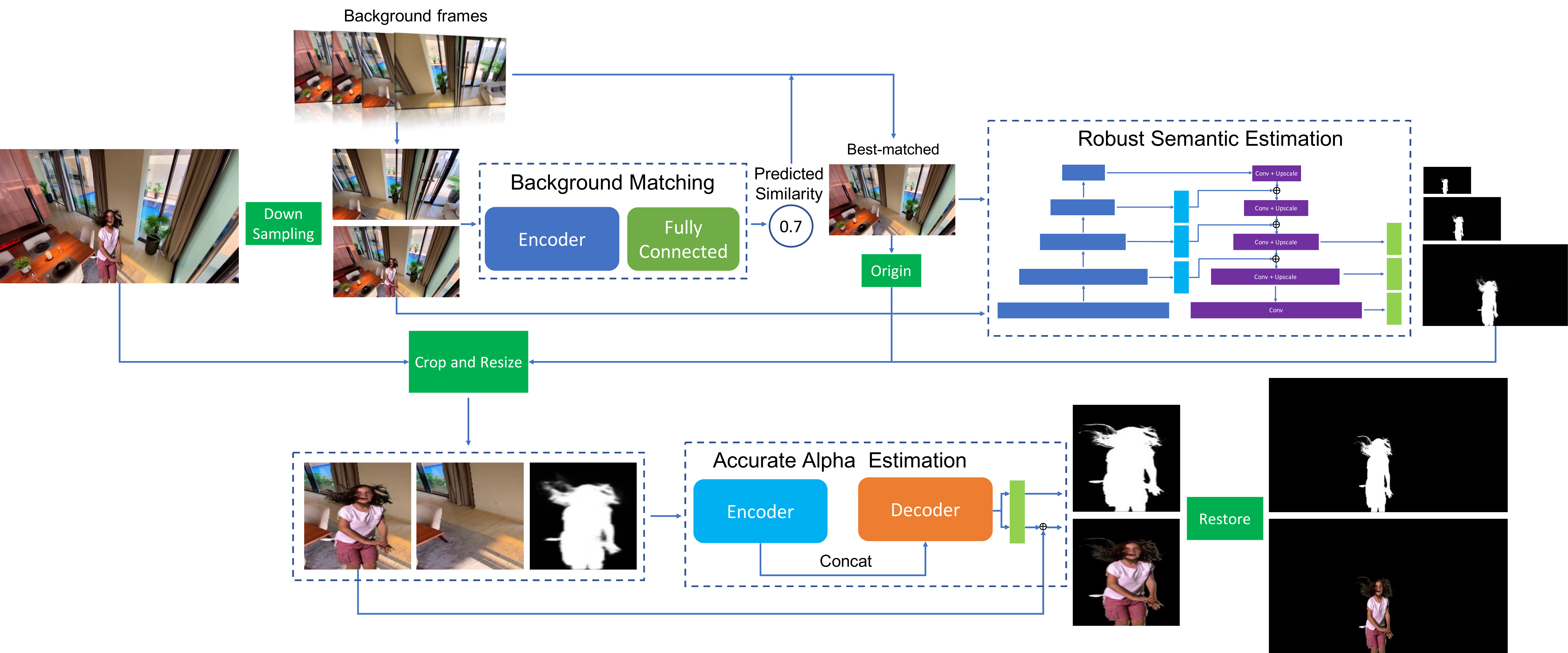} }
  \caption{The pipeline of the proposed method. The proposed method contains three parts: background matching network (BMN), robust semantic estimation network (REN) and accurate alpha estimation network (AEN). BMN aims at finding the most matched background of the current frame from all background frames by estimating the similarity values. In reality, the matched background from BMN may exist misregistration, color and position difference with the current frame. we estimate the alpha matte progressively. REN estimates the coarse alpha matte. With the coarse alpha matte, we remove blank region first and zoom the target area, then estimate the final alpha matte using AEN.}
  \label{fig: flowchart}
\end{figure*}

\section{Proposed Approach}
The proposed method aims at estimating alpha matte using dynamic backgrounds and input videos. There are three parts in the proposed method: background matching network (BMN), robust semantic estimation network (REN) and accurate alpha estimation network (AEN). BMN identifies the corresponding background in the background frames, trying to find the most matched background of the current frame. In reality, the matched background from BMN may exist misregistration, color and position difference with the current frame. Therefore, we estimate the alpha matte progressively, estimating coarse alpha matte first and then refining the coarse alpha matte. REN estimates the coarse alpha matte from the input frame and the matched background from BMN. AEN finally refines the coarse alpha matte using the coarse alpha matte, the input frame and the matched background. The whole flowchart is demonstrated in Figure~\ref{fig: flowchart}.

\subsection{Background Matching Network}
In the proposed method, the background is dynamic and we have to identify the corresponding background with the current input from all background frames. Given the current input image or frame, the target is to find the best-matched background. In real applications, the captured video and dynamic background contain consecutive frames, and the difference among frames is small. Instead of simply classifying the candidate background frame to matched or not matched categories, we try to estimate the similarity value. The background matching network takes two inputs: the input frame, and the background frame. The network outputs the estimated similarity value of the two input images. The structure of BMN is consisted of several convolutional layers, down-sampling layers, and full-connected layers.

During the training process, we have the current frame, the corresponding background, i.e. the ground truth background, and the ground truth alpha matte. We use lots of miss-matched background and simple transformation of the ground truth background to train BMN. We define the similarity value $S$ as one minus the mean $L_1$ difference between the non-foreground region, i.e. the background region,
\begin{equation}
\begin{aligned}
\label{identity_loss}
S(x,y,\alpha)= 1-\frac{\sum_i(|x_i-y_i|_1*(1-\alpha_i))}{\sum_i(1-\alpha_i)},
\end{aligned}
\end{equation}
where $i$ represents the $i$th pixel, $x$ is the current frame, $y$ is the input background frame, $\alpha$ is the corresponding alpha matte, and therefore the non zero elements of $1-\alpha$ indicates the background region.

With the definition of the similarity value, we use the following loss function to train BMN,
\begin{equation}
\begin{aligned}
\label{identity_loss}
\mathcal{L}_{BMN}= |B(x,y)-S(x,y_{gt},\alpha_{gt})|_1,
\end{aligned}
\end{equation}
where $B$ represents the output of BMN, $y_{gt}$ is the ground truth background, $\alpha_{gt}$ is the ground truth alpha matte.


During testing, for all background frames, we use BMN to calculate the similarity value with the current frame. The background frame with the highest similarity value is the best-matched background.

\subsection{Robust Semantic Estimation Network}
Though BEN finds the best-matched background with the current frame from all background frames, it is possible that the found background is not perfectly matched with the current frame. There might exist misregistration, color or position difference in reality. Instead of estimating the alpha matte end-to-end in one step, we choose a progressive way to estimate the alpha matte from coarse to fine. In addition, the foreground region is usually much smaller compared with the large background region in most videos. It is reasonable to estimate a coarse alpha matte first, and then crop and zoom the target foreground region accordingly so as to refine the details. Therefore, in the first step, we use robust semantic estimation network (REN) to estimate the coarse alpha matte.

The structure of REN is well displayed in Figure~\ref{fig: flowchart}, which is basically a feature pyramid network~\cite{lin2017feature}. Low level features are integrated with the high level features through skip connection. Multi-size features are employed to predict the coarse alpha matte with different resolutions during training process. As the target of REN is to estimate the coarse semantic information, we use binary cross entropy loss to train REN.
\begin{equation}
\begin{aligned}
\label{identity_loss}
C(x,y)= -x*log(y)-(1-x)*log(1-y),
\end{aligned}
\end{equation}
where $x$ is the output and $y$ is the ground truth. With multi-scale alpha matte output, we correspondingly calculate multi-scale loss,
\begin{equation}
\begin{aligned}
\label{identity_loss}
\mathcal{L}_{REN}= C(R(x),\alpha_{gt}) + C(R^{\frac{1}{2}}(x),\alpha^{\frac{1}{2}}_{gt}) + C(R^{\frac{1}{4}}(x),\alpha^{\frac{1}{4}}_{gt}),
\end{aligned}
\end{equation}
where $R(x)$ is the full size output of REN, $\alpha_{gt}$ is ground truth alpha matte, $\frac{1}{2}$ and $\frac{1}{4}$ superscript represents the ratio compared with the input size.

\subsection{Accurate Alpha Estimation Network}
We observe that the foreground region is usually a small part of the whole image, and therefore we crop the target foreground region with slight board and zoom it to a fixed large size ($640\times640$ in experiments) to better refine the details, including the input frame, the matched background, and the coarse alpha matte. The cropping operation is based on the coarse alpha matte estimated by REN. Accurate alpha estimation network (AEN) contains encoders and decoders, low level features are concatenated with the high level features so as to predict fine detailed alpha matte. Other than the alpha matte, we simultaneously estimate the foreground residual similarly with~\cite{sengupta2020background, lin2021real}. We use $L_1$ loss to train AEN. The alpha matte loss is,
\begin{equation}
\begin{aligned}
\label{identity_loss}
\mathcal{L}_{\alpha}= |A(x)^{\alpha}-\alpha_{gt}|_1,
\end{aligned}
\end{equation}
The foreground loss is,
\begin{equation}
\begin{aligned}
\label{identity_loss}
\mathcal{L}_{fgr}= |(A(x)^{fgr}+x)*\alpha_{gt}-{fgr}_{gt}*\alpha_{gt}|_1,
\end{aligned}
\end{equation}
where $A(x)^{\alpha}$ is the output alpha matte, $A(x)^{fgr}$ is the output foreground residual, $\alpha_{gt}$ is the ground truth alpha matte,$fgr_{gt}$ is the ground truth foreground. The final loss function is the sum of alpha loss and foreground loss.

\begin{equation}
\begin{aligned}
\label{identity_loss}
\mathcal{L}_{AEN}= \mathcal{L}_{\alpha}+\mathcal{L}_{fgr}.
\end{aligned}
\end{equation}

\subsection{Implementation details}
\paragraph{Training details.} For speed consideration, both BMN and AEN are trained at a low resolution. The input and the background are downscaled to $192\times320$. When training AEN, all images are cropped according to the coarse alpha matte and then zoomed to $640\times640$. BMN is trained independently with BMN and AEN. REN and AEN are trained respectively for some epochs and then we co-train REN and AEN.
\paragraph{Background matching acceleration.} The proposed method manages to find the best-matched background from all background frames by estimating the similarity of the current frame and all candidate background frames. It can be very time-consuming when the background video is long. Normally, the background videos contain consecutive frames, and the difference among neighboring frames is small. Instead of finding the best matched background from all background frames, we reduce the candidate background frames first using fixed interval sampling. For example, if we set the sampling interval to 8, only 12.5\% of the background frames will be used, accelerating the background matching process significantly.

\begin{figure*}[ht]
  \centering
  \resizebox{1\linewidth}{!}{
   \includegraphics{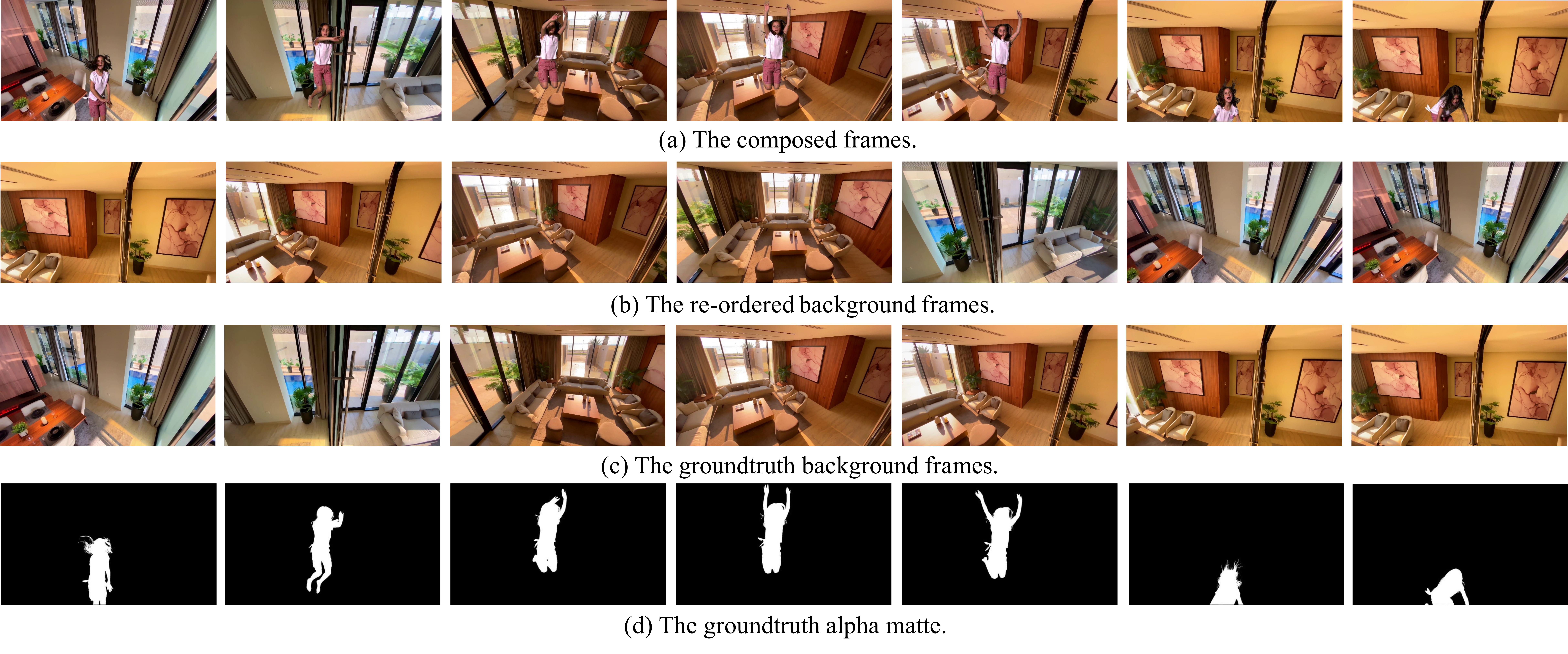} }
  \caption{The dynamic background matting dataset. Both the foreground and background are dynamic. The background frame miss-matches the composed frame frame-wisely. (a) and (b) well simulate the user capturing the video freely and then capturing the background regions randomly in reality.}
  \label{fig: dataset}
\end{figure*}

\section{Dynamic Background Matting Dataset}
Different from previous works, we aims to solve the dynamic background problem, requiring the background to change frame by frame. To be more realistic, the foreground object should move frame by frame at the same time, to simulate the scene captured by a moving camera. VideoMatte240K~\cite{lin2021real} dataset contains 482 high quality videos, with groundtruth alpha matte and foreground. We collect 102 dynamic videos from the internet and manually compose the videos from VideoMatte240K with the 102 collected dynamic videos to create our dynamic background dataset. In our dataset, both the foreground and the background change frame by frame. After composing the videos, we reverse the background videos to serve as the captured backgrounds videos. In reality, the captured background is random, and not matched with the freely captured video frame by frame. This reversion operation ensures that the captured background video miss-matches the composed video frame-wisely.

478 foreground videos from VideoMatte240K dataset, coupled with 82 collected dynamic videos, compose our training dataset. Each video is combined with 20 background videos, making up 9560 dynamic videos. The left 4 foreground videos and 12 dynamic background videos are used to create the testing dataset, making up 48 testing videos, 22933 testing images. There is no overlap between the training and testing dataset regarding to both the foreground videos and the background videos. Some samples of our dataset is displayed in Figure~\ref{fig: dataset}.

\begin{figure*}[ht]
  \centering
  \resizebox{1\linewidth}{!}{
   \includegraphics{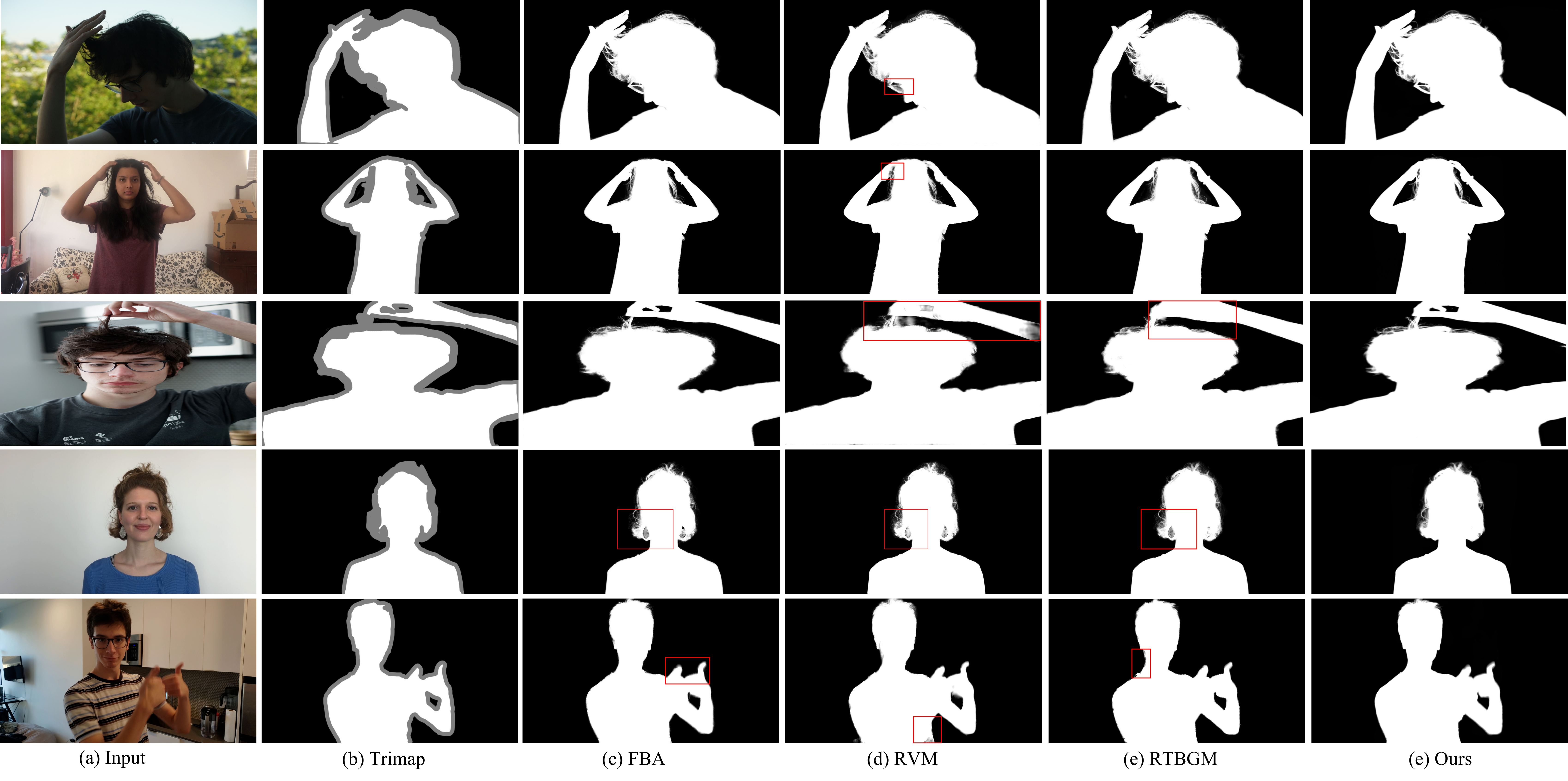} }
  \caption{Qualitative evaluation on real images from~\cite{lin2021real}. RVM~\cite{lin2022robust} suffers from inaccurate semantic estimation due to the lack of large training dataset. FBA~\cite{forte2020f}, RTBGM~\cite{lin2021real} and the proposed perform stably and output excellent hair details. The proposed method performs better when the foreground is close to the background.}
  \label{fig: compare}
\end{figure*}z

\section{Experiments}
\subsection{Evaluation results}
\paragraph{Evaluation metrics.} The most widely used metrics to measure the matting results are mean square error (MSE), sum of absolution difference (SAD), gradient and connectivity error~\cite{xu2017deep,chen2018semantic}. The gradient and connectivity is able to measure the quality of alpha matte similarly to human perception~\cite{rhemann2009perceptually}. We calculate all four four metrics for quantitative comparison purpose. As non-trimap-based methods need to stimate per-pixel alpha matte, we calculate all metrics using the whole image.
\paragraph{Baselines.} We categorize matting methods to three categories: trimap-based methods, trimap-free methods, and background-based methods. Therefore, we compare the proposed method with all three kinds methods, FBA~\cite{forte2020f}, RVM~\cite{lin2022robust} and RTBGM~\cite{lin2021real}. FBA is the representative of trimap-based methods. RVM is one of the latest trimap-free video matting methods. RTBGM represents the latest background-based methods. As RTBGM requires static background, we test RTBGM using ground truth backgrounds.
\paragraph{Performance comparison.} We compare all fours methods quantitatively on synthetic testing dataset and qualitatively using real images from ~\cite{lin2021real}. Using the synthetic testing dataset, we calculate the evaluation metrics corresponding to all four methods (Table~\ref{tab:compare}). FBA performs the best with the given GT trimap. Though GT background is used for RTBGM and the proposed method needs to identify proper background from video frames, the metrics of RTBGM and the proposed method is close, all of which are significantly better than trimap-free method RVM. In Figure~\ref{fig: compare}, we display the qualitative results on real images. The trimap for FBA is generated manually using Photoshop. Generally, FBA, RTBGM and the proposed method perform stably on all images and output excellent hair details. Trimap-free method RVM suffers from inaccurate semantic estimation. Trimap-free methods has no extra assistance and require large amount of training dataset so as to estimate accurate semantic information. With respect to the scene when the foreground is close to the background, FBA and RTBGM wrongly segment part of foreground pixels to background. In comparison, the proposed method output accurate alpha matte.
\paragraph{Running time.} All methods are tested using 720P images, Nvidia P100 graphic card. The running time is listed in Table~\ref{tab:compare}. FBA~\cite{forte2020f} takes 377$ms$, RVM~\cite{lin2022robust} takes about 12.1$ms$, RTBGM~\cite{lin2021real} takes 24.1$ms$. For our method, the matting time (REN+AEN) is 34.8ms, which is also very fast. Note that here we only compare the matting time of the four methods. In reality, the background matching network takes 2.83ms per-image to estimate the similarly value. So it will take another $2.83n\ ms$ to find the best matched ground from $n$ background frames.


\begin{table}[t]
  \centering\scriptsize
    \caption{The quantitative results on testing dataset.}
  \resizebox{1\linewidth}{!}{
  \begin{tabular}{lcccc}
    \hline
    Method &FBA~\cite{forte2020f}  &RVM~\cite{lin2022robust}  &RTBGM~\cite{lin2021real}  &ours   \\
    \hline
    SAD($1e^{-3}$) & 4.04 & 5.84 & 4.46 & 4.45 \\
    MSE($1e^{-3}$) & 0.285 & 1.25 & 0.467 & 0.362 \\
    Gradient($1e^{-3}$) & 0.304 & 0.629 & 0.320 & 0.395 \\
    Connectivity($1e^{-3}$) & 0.979 & 2.623 & 1.19 & 1.23 \\
    \hline
    Running time(ms) & 377 & 12.1 & 24.1 & 34.8 (REN+AEN)\\
    \hline
    \end{tabular}}
  \label{tab:compare}
\end{table}

\begin{figure*}[ht]
  \centering
  \resizebox{1\linewidth}{!}{
   \includegraphics{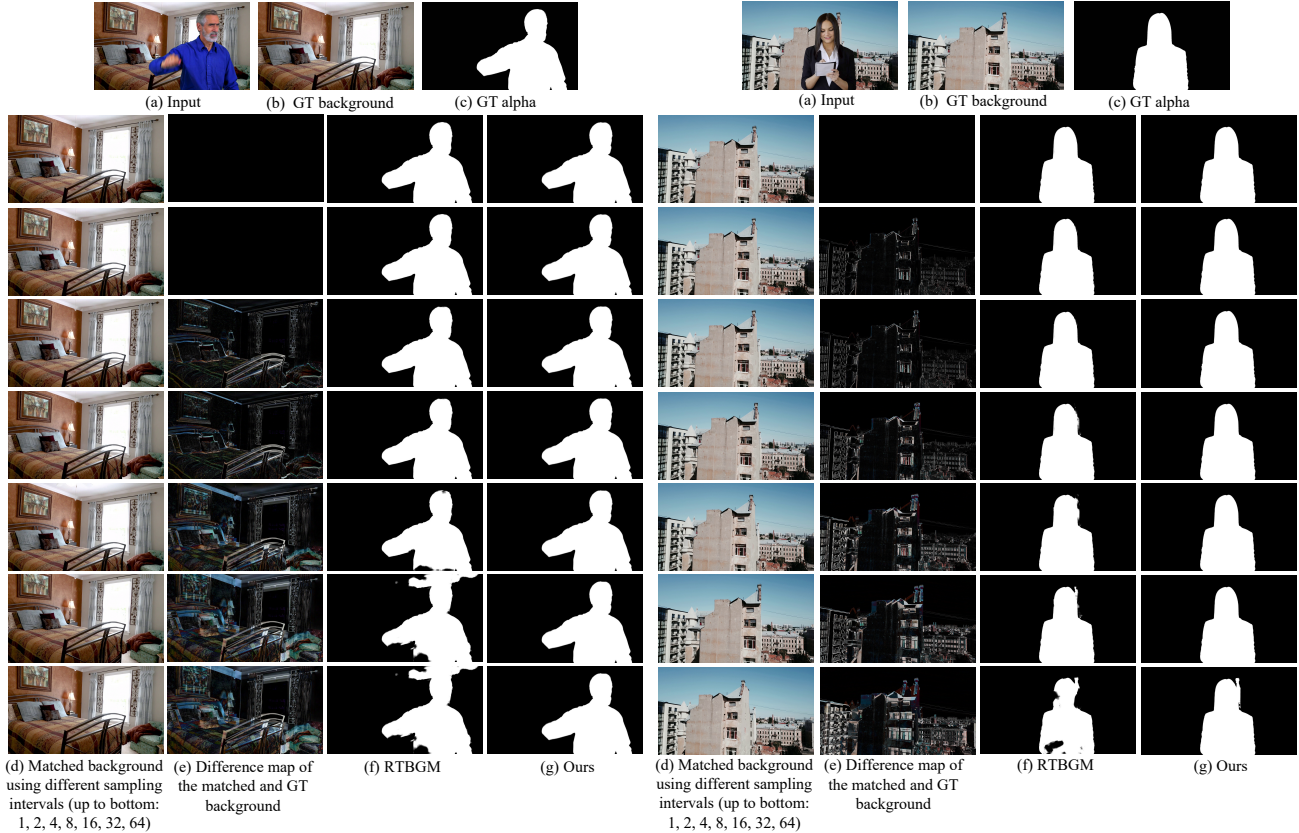} }
  \caption{Comparison of RTBGM~\cite{lin2021real} and the proposed method taking in the same background with respect to different sampling intervals (background accuracy). The difference between the best-matched background and the groundtruth backgrounds increases with the growth of sampling intervals (from up to bottom, the sampling interval is 1, 2, 4, 8, 16, 32, 64 respectively). Using the same matched background, the proposed method performs more stably than RTBGM. The performance of RTBGM becomes worse significantly when the sampling interval is greater than 8 (the fifth row). Whereas, the proposed method performs stably even when the sampling interval is 32 (the second row from the bottom).}
  \label{fig: sampling_interval}
\end{figure*}

\begin{table*}[]
    \centering\scriptsize
      \caption{The quantitative results when using different background sampling intervals.}
    \resizebox{1\linewidth}{!}{
\begin{tabular}{ll|cccccccc}
  \hline
\multicolumn{2}{l|}{Background sampling interval} &1  &2  &4  &8  &16  &32  &64  \\ \hline
\multicolumn{2}{l|}{Difference between the matched and GT background} &2.13  &4.40  &6.87  &10.04  &13.71  &18.01  &23.88   \\ \hline
\multicolumn{1}{l|}{\multirow{4}{*}{Matched bk + RTBGM~\cite{lin2021real}}} & SAD($1e^{-3}$) &4.52  &4.92  &5.54  &7.27  &11.08  &24.09  &66.65  \\
\multicolumn{1}{l|}{} & MSE($1e^{-3}$) &0.499  &0.849  &1.41  &3.03  &6.63  &19.2  &61.1    \\
\multicolumn{1}{l|}{} & Gradient($1e^{-3}$) &0.336  &0.376  &0.420  &0.496  &0.609  &0.807  &1.15    \\
\multicolumn{1}{l|}{} & Connectivity($1e^{-3}$) &1.23  &1.98  &2.12  &3.93  &7.69  &21.1  &63.8    \\
\hline
\multicolumn{1}{l|}{\multirow{4}{*}{Matched bk + Ours}}  & SAD($1e^{-3}$)   &4.45  &4.48  &4.51  &4.63  &4.98  &6.12  &12.3    \\
\multicolumn{1}{l|}{} & MSE($1e^{-3}$)   &0.362  &0.382  &0.411  &0.520  &0.833  &1.93  &7.92    \\
\multicolumn{1}{l|}{} & Gradient($1e^{-3}$)   &0.395  &0.401  &0.407  &0.419  &0.439  &0.476  &0.580    \\
\multicolumn{1}{l|}{} & Connectivity($1e^{-3}$)   &1.23  &1.27  &1.29  &1.42  &1.77  &2.95  &8.97   \\
\hline
\end{tabular}}
\label{tab:sampling_interval}
\end{table*}

\paragraph{Ablation study.} The proposed method is composed of background matching and background matting. To measure the quality of the proposed background matting method, we compare RTBGM and the proposed using the same input background, i.e. the matched background. In the proposed method, we propose to reduce the number of candidate background frames by fixed interval sampling to accelerate the background matching process. Considering lots of frames being removed by the sampling strategy, it is possible that the found best-matched background is not perfectly matched, existing some difference with the input frame. This misregistration may impact the final performance. Therefore, we also test the performance of the proposed method when the matched background exists misregistration with the input frame.

In Figure~\ref{fig: sampling_interval}, we display the results of RTBGM and the proposed method taking in the same background with respect to different sampling intervals. In addition, we draw the difference map between the best-matched background and the groundtruth background. From the difference map, we can observe that when the sampling interval is large, the found best matched background shows larger difference with the groundtruth background. Even so, the proposed method performs consistently well regarding to different sampling intervals, benefiting from our robust semantic estimation network. In comparison, RTBGM fails to perform well when the difference is large. In Table~\ref{tab:sampling_interval}, we calculate all four metrics to compare the results quantitatively. We also calculate mean $L_1$ difference between the matched background and the groundtruth background. The difference increases with the growth of sampling intervals. The performance of RTBGM becomes worse significantly when the sampling interval is greater than 8. In contrast, the proposed method performs stably when the sampling interval is less than 32.

\subsection{Applying to real world video}
\begin{figure}[t]
  \centering
  \resizebox{1\linewidth}{!}{
   \includegraphics{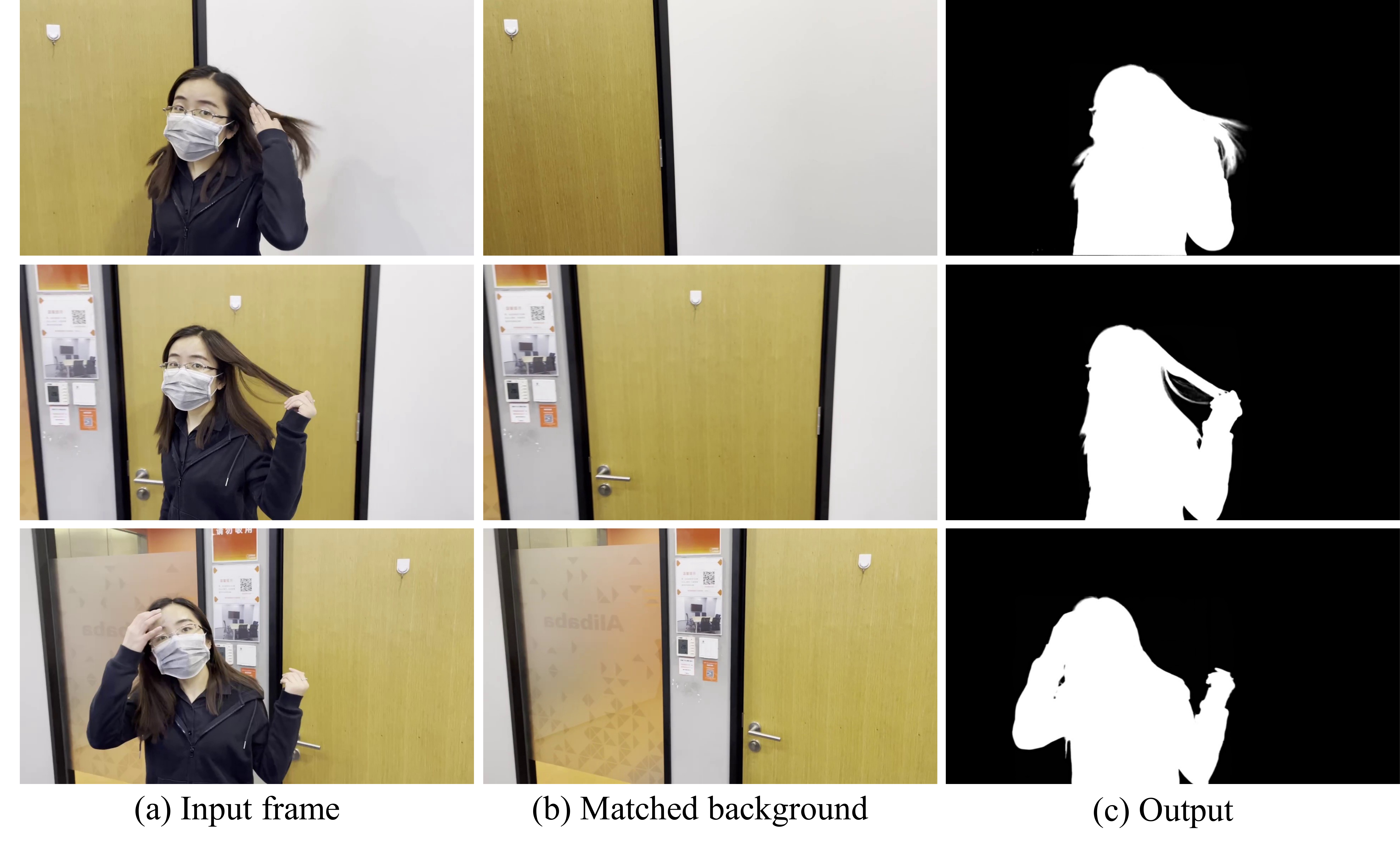} }
  \caption{Applying to real world video. For fast background matting, the sampling interval is set to 8. The proposed method is able to find roughly matched background from background frames. Though the best-matched background exhibits some misregistration with the input frame, the proposed method outputs promising results.}
  \label{fig: real_world}
\end{figure}

The greatest advantage of the proposed method is that the user now is not limited to capture static background, which makes it much easier to use in real world applications than previous background-based works. The user first captures their videos freely, moving the cameras as they want. Then the user captures the background, roughly covering the background regions. In Figure~\ref{fig: real_world}, we demonstrate the results on real world videos. The background video lasts for 12s, containing 344 frames. The sampling interval is set to 8. For background matting process, we search from $344/8=43$ frames to find the best matched background. The full inference time is $2.83\times43+34.8=156ms$ per-frame. The proposed method is able to find roughly matched background from background frames. Though the best-matched background exhibits some misregistration with the input frame, the results are very promising.

Also, we conduct user study using 35 real videos, comparing with RTBGM. The proposed method is proved to be robust with background variations. Whereas RTBGM is designed for static background and is sensitive to background variations (Figure 5). Our method predominates the user study in experiments. 81\% participants consider that the results of the proposed method are better than RTBGM.

\section{Limitations and Discussions}
The first step of the proposed method is to find the matched background frame from the captured background video. Using Nvidia P100 graphic card, the inference time of background matching network is about 2.83ms. Even so, it can be time-consuming when the captured background video is long. If we set the background sampling interval to 8, it takes seconds if we have to search for the best matched background from more than 2826 frames (about 112s long video). For applications where the running time is important, we recommend to take background videos no more than 100 seconds. Whereas, if the running time is not the limit, long videos are also applicable.

In the proposed pipeline, the user first captures normal videos, then the background. There might be a consideration whether the users have to take the background very carefully, trying to duplicate the video capturing process. As we have shown in Figure~\ref{fig: sampling_interval}, the proposed method is robust against background misregistration. Therefore, the user only needs to cover the background region roughly.

\section{Conclusion}
The limitations of background-based methods motivate us to propose a dynamic background matting method. To achieve this goal, we use background matching network to find the best-matched background frame by frame from dynamic backgrounds. For acceleration purpose, we use fixed interval sampling to reduce the background matching time. Then, robust semantic estimation network is used to estimate the coarse alpha matte. Finally, we crop and zoom the target region according to the coarse alpha matte, and estimate the final accurate alpha matte. The proposed method performs stably and comparably against the state-of-the-art methods. The proposed method allows the user to move their camera when capturing videos in real applications. Considering the limitations of static background matting methods (not flexible), green screen matting (not convenient), trimap-based (not convenient) and trimap-free (not stable) matting methods, the proposed method (stable, flexible and convenient) probably reaches the best balance between convenience and matting quality.

{\small
\bibliographystyle{ieee_fullname}
\bibliography{egbib}

\begin{thebibliography}{10}\itemsep=-1pt

\bibitem{aksoy2018semantic}
Ya{\u{g}}iz Aksoy, Tae-Hyun Oh, Sylvain Paris, Marc Pollefeys, and Wojciech
  Matusik.
\newblock Semantic soft segmentation.
\newblock {\em ACM Transactions on Graphics (TOG)}, 37(4):72, 2018.

\bibitem{aksoy2017designing}
Yagiz Aksoy, Tunc Ozan~Aydin, and Marc Pollefeys.
\newblock Designing effective inter-pixel information flow for natural image
  matting.
\newblock In {\em The IEEE Conference on Computer Vision and Pattern
  Recognition (CVPR)}, pages 29--37. IEEE, 2017.

\bibitem{bai2007geodesic}
Xue Bai and Guillermo Sapiro.
\newblock A geodesic framework for fast interactive image and video
  segmentation and matting.
\newblock In {\em The IEEE International Conference on Computer Vision (ICCV)},
  pages 1--8. IEEE, 2007.

\bibitem{Cai_2019_ICCV}
Shaofan Cai, Xiaoshuai Zhang, Haoqiang Fan, Haibin Huang, Jiangyu Liu, Jiaming
  Liu, Jiaying Liu, Jue Wang, and Jian Sun.
\newblock Disentangled image matting.
\newblock In {\em The IEEE International Conference on Computer Vision (ICCV)}.
  IEEE, 2019.

\bibitem{chen2018semantic}
Quan Chen, Tiezheng Ge, Yanyu Xu, Zhiqiang Zhang, Xinxin Yang, and Kun Gai.
\newblock Semantic human matting.
\newblock In {\em Proceedings of the 26th ACM international conference on
  Multimedia}, pages 618--626. ACM, 2018.

\bibitem{chen2013knn}
Qifeng Chen, Dingzeyu Li, and Chi-Keung Tang.
\newblock Knn matting.
\newblock {\em IEEE Transactions on Pattern Analysis and Machine Intelligence
  (TPAMI)}, 35(9):2175--2188, 2013.

\bibitem{chuang2001bayesian}
Yung-Yu Chuang, Brian Curless, David~H Salesin, and Richard Szeliski.
\newblock A bayesian approach to digital matting.
\newblock In {\em The IEEE Conference on Computer Vision and Pattern
  Recognition (CVPR)}, pages 264--271. IEEE, 2001.

\bibitem{feng2016cluster}
Xiaoxue Feng, Xiaohui Liang, and Zili Zhang.
\newblock A cluster sampling method for image matting via sparse coding.
\newblock In {\em The European Conference on Computer Vision (ECCV)}, pages
  204--219. Springer, 2016.

\bibitem{forte2020f}
Marco Forte and Fran{\c{c}}ois Piti{\'e}.
\newblock $ f $, $ b $, alpha matting.
\newblock {\em arXiv preprint arXiv:2003.07711}, 2020.

\bibitem{gastal2010shared}
Eduardo~SL Gastal and Manuel~M Oliveira.
\newblock Shared sampling for real-time alpha matting.
\newblock In {\em Computer Graphics Forum}, volume~29, pages 575--584. Wiley
  Online Library, 2010.

\bibitem{grady2005random}
Leo Grady, Thomas Schiwietz, Shmuel Aharon, and R{\"u}diger Westermann.
\newblock Random walks for interactive alpha-matting.
\newblock In {\em Proceedings of VIIP}, volume 2005, pages 423--429, 2005.

\bibitem{he2011global}
Kaiming He, Christoph Rhemann, Carsten Rother, Xiaoou Tang, and Jian Sun.
\newblock A global sampling method for alpha matting.
\newblock In {\em The IEEE Conference on Computer Vision and Pattern
  Recognition (CVPR)}, pages 2049--2056. IEEE, 2011.

\bibitem{hou2019context}
Qiqi Hou and Feng Liu.
\newblock Context-aware image matting for simultaneous foreground and alpha
  estimation.
\newblock In {\em Proceedings of the IEEE/CVF International Conference on
  Computer Vision}, pages 4130--4139, 2019.

\bibitem{Hou_2019_ICCV}
Qiqi Hou and Feng Liu.
\newblock Context-aware image matting for simultaneous foreground and alpha
  estimation.
\newblock In {\em The IEEE International Conference on Computer Vision (ICCV)}.
  IEEE, 2019.

\bibitem{johnson2016sparse}
Jubin Johnson, Ehsan~Shahrian Varnousfaderani, Hisham Cholakkal, and Deepu
  Rajan.
\newblock Sparse coding for alpha matting.
\newblock {\em IEEE Transactions on Image Processing (TIP)}, 25(7):3032--3043,
  2016.

\bibitem{karacan2015image}
Levent Karacan, Aykut Erdem, and Erkut Erdem.
\newblock Image matting with kl-divergence based sparse sampling.
\newblock In {\em The IEEE International Conference on Computer Vision (ICCV)},
  pages 424--432. IEEE, 2015.

\bibitem{levin2007closed}
Anat Levin, Dani Lischinski, and Yair Weiss.
\newblock A closed-form solution to natural image matting.
\newblock {\em IEEE Transactions on Pattern Analysis and Machine Intelligence
  (TPAMI)}, 30(2):228--242, 2007.

\bibitem{levin2008spectral}
Anat Levin, Alex Rav-Acha, and Dani Lischinski.
\newblock Spectral matting.
\newblock {\em IEEE Transactions on Pattern Analysis and Machine Intelligence
  (TPAMI)}, 30(10):1699--1712, 2008.

\bibitem{lin2021real}
Shanchuan Lin, Andrey Ryabtsev, Soumyadip Sengupta, Brian~L Curless, Steven~M
  Seitz, and Ira Kemelmacher-Shlizerman.
\newblock Real-time high-resolution background matting.
\newblock In {\em Proceedings of the IEEE/CVF Conference on Computer Vision and
  Pattern Recognition}, pages 8762--8771, 2021.

\bibitem{lin2022robust}
Shanchuan Lin, Linjie Yang, Imran Saleemi, and Soumyadip Sengupta.
\newblock Robust high-resolution video matting with temporal guidance.
\newblock In {\em Proceedings of the IEEE/CVF Winter Conference on Applications
  of Computer Vision}, pages 238--247, 2022.

\bibitem{lin2017feature}
Tsung-Yi Lin, Piotr Doll{\'a}r, Ross Girshick, Kaiming He, Bharath Hariharan,
  and Serge Belongie.
\newblock Feature pyramid networks for object detection.
\newblock In {\em Proceedings of the IEEE conference on computer vision and
  pattern recognition}, pages 2117--2125, 2017.

\bibitem{liu2020boosting}
Jinlin Liu, Yuan Yao, Wendi Hou, Miaomiao Cui, Xuansong Xie, Changshui Zhang,
  and Xian-sheng Hua.
\newblock Boosting semantic human matting with coarse annotations.
\newblock In {\em Proceedings of the IEEE/CVF Conference on Computer Vision and
  Pattern Recognition}, pages 8563--8572, 2020.

\bibitem{lu2019indices}
Hao Lu, Yutong Dai, Chunhua Shen, and Songcen Xu.
\newblock Indices matter: Learning to index for deep image matting.
\newblock In {\em Proceedings of the IEEE/CVF International Conference on
  Computer Vision}, pages 3266--3275, 2019.

\bibitem{qiao2020attention}
Yu Qiao, Yuhao Liu, Xin Yang, Dongsheng Zhou, Mingliang Xu, Qiang Zhang, and
  Xiaopeng Wei.
\newblock Attention-guided hierarchical structure aggregation for image
  matting.
\newblock In {\em Proceedings of the IEEE/CVF Conference on Computer Vision and
  Pattern Recognition}, pages 13676--13685, 2020.

\bibitem{rhemann2009perceptually}
Christoph Rhemann, Carsten Rother, Jue Wang, Margrit Gelautz, Pushmeet Kohli,
  and Pamela Rott.
\newblock A perceptually motivated online benchmark for image matting.
\newblock In {\em 2009 IEEE Conference on Computer Vision and Pattern
  Recognition}, pages 1826--1833. IEEE, 2009.

\bibitem{ruzon2000alpha}
Mark~A Ruzon and Carlo Tomasi.
\newblock Alpha estimation in natural images.
\newblock In {\em The IEEE Conference on Computer Vision and Pattern
  Recognition (CVPR)}, volume~1, pages 18--25. IEEE, 2000.

\bibitem{sengupta2020background}
Soumyadip Sengupta, Vivek Jayaram, Brian Curless, Steven~M Seitz, and Ira
  Kemelmacher-Shlizerman.
\newblock Background matting: The world is your green screen.
\newblock In {\em Proceedings of the IEEE/CVF Conference on Computer Vision and
  Pattern Recognition}, pages 2291--2300, 2020.

\bibitem{shen2016deep}
Xiaoyong Shen, Xin Tao, Hongyun Gao, Chao Zhou, and Jiaya Jia.
\newblock Deep automatic portrait matting.
\newblock In {\em The European Conference on Computer Vision (ECCV)}, pages
  92--107. Springer, 2016.

\bibitem{sun2004poisson}
Jian Sun, Jiaya Jia, Chi-Keung Tang, and Heung-Yeung Shum.
\newblock Poisson matting.
\newblock In {\em ACM Transactions on Graphics (ToG)}, volume~23, pages
  315--321. ACM, 2004.

\bibitem{sun2021semantic}
Yanan Sun, Chi-Keung Tang, and Yu-Wing Tai.
\newblock Semantic image matting.
\newblock In {\em Proceedings of the IEEE/CVF Conference on Computer Vision and
  Pattern Recognition}, pages 11120--11129, 2021.

\bibitem{tang2019learning}
Jingwei Tang, Yagiz Aksoy, Cengiz Oztireli, Markus Gross, and Tunc~Ozan Aydin.
\newblock Learning-based sampling for natural image matting.
\newblock In {\em Proceedings of the IEEE/CVF Conference on Computer Vision and
  Pattern Recognition}, pages 3055--3063, 2019.

\bibitem{wang2008image}
Jue Wang, Michael~F Cohen, et~al.
\newblock Image and video matting: a survey.
\newblock {\em Foundations and Trends{\textregistered} in Computer Graphics and
  Vision}, 3(2):97--175, 2008.

\bibitem{xu2017deep}
Ning Xu, Brian Price, Scott Cohen, and Thomas Huang.
\newblock Deep image matting.
\newblock In {\em Proceedings of the IEEE Conference on Computer Vision and
  Pattern Recognition}, pages 2970--2979, 2017.

\bibitem{Zhang_2019_CVPR}
Yunke Zhang, Lixue Gong, Lubin Fan, Peiran Ren, Qixing Huang, Hujun Bao, and
  Weiwei Xu.
\newblock A late fusion cnn for digital matting.
\newblock In {\em The IEEE Conference on Computer Vision and Pattern
  Recognition (CVPR)}. IEEE, 2019.

\bibitem{zhu2017fast}
Bingke Zhu, Yingying Chen, Jinqiao Wang, Si Liu, Bo Zhang, and Ming Tang.
\newblock Fast deep matting for portrait animation on mobile phone.
\newblock In {\em Proceedings of the 25th ACM international conference on
  Multimedia}, pages 297--305. ACM, 2017.

\end{thebibliography}
}

\end{document}